\documentclass{article}

\usepackage{arxiv}

\usepackage[utf8]{inputenc} 
\usepackage[T1]{fontenc}    
\usepackage{hyperref}       
\usepackage{url}            
\usepackage{booktabs}       
\usepackage{amsfonts}       
\usepackage{nicefrac}       
\usepackage{microtype}      
\usepackage{lipsum}		
\usepackage{graphicx}
\usepackage{doi}
\usepackage{adjustbox}
\usepackage{multirow}
\usepackage{caption}
\usepackage{float}
\usepackage{array}
\usepackage[numbers]{natbib}

\title{Generative adversarial networks vs large language models: a comparative study on synthetic tabular data generation}

\author{
    Austin A. Barr\thanks{Correspondence: Austin A. Barr, austin.barr@ucalgary.ca} \\
	Cumming School of Medicine\\
	University of Calgary\\
	Calgary, AB, Canada\\
    \And
    Robert Rozman \\
    Independent Researcher \\
    Toronto, ON, Canada \\
	\And
	Eddie Guo \\
	Cumming School of Medicine\\
	University of Calgary\\
	Calgary, AB, Canada\\
}

\date{}

\renewcommand{\undertitle}{}
\renewcommand{\shorttitle}{CTGAN vs LLMs on Synthetic Tabular Data}

\hypersetup{
pdftitle={CTGAN vs LLMs on Synthetic Tabular Data},
pdfsubject={cs.CL},
pdfauthor={Austin A. Barr, Robert Rozman, Eddie Guo},
pdfkeywords={Synthetic Data, Large Language Models, Generative Adversarial Networks, Artificial Intelligence },
}

\begin{document}
\maketitle

\begin{abstract}
We propose a new framework for zero-shot generation of synthetic tabular data. Using the large language model (LLM) GPT-4o and plain-language prompting, we demonstrate the ability to generate high-fidelity tabular data without task-specific fine-tuning or access to real-world data (RWD) for pre-training. To benchmark GPT-4o, we compared the fidelity and privacy of LLM-generated synthetic data against data generated with the conditional tabular generative  adversarial network (CTGAN), across three open-access datasets: \textit{Iris}, \textit{Fish Measurements}, and \textit{Real Estate Valuation}. Despite the zero-shot approach, GPT-4o outperformed CTGAN in preserving means, 95\% confidence intervals, bivariate correlations, and data privacy of RWD, even at amplified sample sizes. Notably, correlations between parameters were consistently preserved with appropriate direction and strength. However, refinement is necessary to better retain distributional characteristics. These findings highlight the potential of LLMs in tabular data synthesis, offering an accessible alternative to generative adversarial networks and variational autoencoders.
\end{abstract}


\section{Introduction}
Data-driven research often requires access to large, high-quality datasets. However, real-world data (RWD) is frequently scarce, incomplete, unstructured, or non-standardized, which constrains its use for research \cite{Newgard2015, Fan2014}. Datasets may also be sensitive, and cannot be readily accessed, shared, and used without significant, yet necessary, regulatory hurdles (i.e., research ethics board approval, data-sharing agreements, data de-identification) \cite{Pavlenko2020, LQ2020, TS2017}. Increasingly, synthetic data is positioned as a solution to overcome these limitations \cite{Jordon2022, Bellovin2018, FR2022}.\\
\\
Synthetic data is artificial data which is generated to retain statistical properties of RWD while aiming to preserve data privacy. Synthetic tabular data has applicability toward enhancing RWD (i.e., amplification, augmentation), creating new training samples for machine learning (ML) models, and enabling further data sharing \cite{FR2022}. Despite these promising applications, generating high-quality synthetic data remains a challenging task. Synthetic tabular data requires preserving dataset structures, domain-specific constraints, means, proportions, distributions, and relationships between features. \\
\\
To evaluate the quality of synthetic data, three primary metrics are used: fidelity, utility, and privacy \cite{Jordon2022}. Fidelity and utility are closely related; fidelity measures the statistical alignment between synthetic and RWD, while utility assesses whether synthetic data can serve as a viable substitute in downstream tasks, such as training ML models. Privacy considerations are also important, as synthetic data should minimize risks of revealing sensitive information from RWD. This presents a privacy-utility trade-off \cite{FR2022}, where enhancing fidelity and utility may increase susceptibility to privacy attacks and re-identification. Conversely, techniques such as differential privacy—which introduce controlled noise to obscure individual records—may reduce the utility of synthetic data \cite{Jordon2018}.\\
\\
In this paper, we investigate the potential for zero-shot generation of synthetic tabular data using large language models (LLMs). To this end, four prominent LLMs—GPT-4o (OpenAI), Claude 3.5 Sonnet (Anthropic), Gemini Advanced 1.5 Pro (Google), and Le Chat (Mistral)—are evaluated on the ability to generate tabular data without fine-tuning or access to RWD for pre-training. The LLM-generated data is benchmarked using data generated with the conditional tabular generative adversarial network (CTGAN) \cite{Xu2019}, across three open-access datasets: \textit{Iris}, \textit{Fish Measurements}, and \textit{Real Estate Valuation}. Using several validated metrics, assessments of data fidelity and privacy are performed. 

\section{Related Work}
\subsection{Generative Adversarial Networks}
Generative adversarial networks (GANs) \cite{Goodfellow2014} have been effectively applied to synthesize various forms of synthetic data \cite{Figueira2022, Yi2019}. In the context of tabular data, CTGAN is a GAN-based model specifically designed for tabular data generation \cite{Xu2019}. A considerable body of literature has demonstrated the fidelity and utility of data generated with CTGAN, across several applications and disciplines \cite{Xu2019, Figueira2022, Kang2023, Mendikowski2022}. However, like other GAN-based approaches, CTGAN requires access to RWD, which presents privacy concerns. Overfitting and memorization of training data increases susceptibility to privacy attacks and data leak \cite{Chen2020, Hayes2017}. This vulnerability is particularly of concern when GANs are trained on RWD with smaller sample sizes \cite{Chen2020}. GAN-based models also require some degree of technical expertise to generate high-quality synthetic data (e.g., GAN architecture, optimization, fine-tuning), which poses additional limitations to the accessibility of this approach. 
\subsection{Variational Autoencoders}
Variational autoencoders (VAEs) are another class of deep generative models used to synthesize synthetic data \cite{Kingma2013}. Several VAEs have demonstrated applicability toward tabular data generation, including TVAE \cite{Xu2019}, which was presented alongside CTGAN. Like GAN-based approaches, VAEs require access to RWD and technical expertise.
\subsection{Large Language Models}
LLMs are another form of generative artificial intelligence, which have been applied to synthetic data generation. This primarily includes text-based data, including synthetic interview transcripts \cite{Ham2023} and medical records \cite{CL2024}. Some approaches to LLM-based tabular data synthesis have also emerged, including GReaT \cite{Borisov2022} and TabuLa \cite{Zhao2023}. However, current LLM-based frameworks require access to RWD, pre-training, and/or fine-tuning. More recently, preliminary evidence has demonstrated the potential for zero-shot generation of synthetic tabular data with the LLM, GPT-4o \cite{BarrFAI2025, BarrarXiv2025}. Without access to RWD, GPT-4o was capable of generating tabular data with high fidelity to clinical RWD by using plain-language prompts which described the desired statistical properties. Notably, this initial evidence has demonstrated the preservation of relationships between parameters, synthesis of new and interrelated features, amplification of sample sizes, and the utility of LLM-generated data toward training ML models. However, it remains unclear whether observed fidelity would scale to further interactions between parameters and how performance would compare to GAN-based frameworks with further benchmarking metrics. The potential for tabular data generation without requiring technical expertise and access to RWD in a zero-shot setting warrants further investigation. 

\section{Methods}
\subsection{Datasets and Preparation}
Three datasets were used in the present analysis, described in Table \ref{tab:datasets}. Two of the datasets were sourced from the University of California, Irvine Machine Learning Repository, and a third from Kaggle. The datasets were pre-processed to ensure completeness and to remove non-numeric columns. No outlier removal was performed. Each dataset’s univariate (mean, standard deviation, range) and bivariate (Pearson’s product moment correlation) statistical properties were summarized to inform the prompts for data generation.

\begin{table}[h]
    \captionsetup{width=0.8\linewidth}
    \caption{Summary of open-access real-world datasets.}
    \vspace{10pt}
	\centering
    \centering
    \begin{adjustbox}{max width=1\textwidth,center}
    \begin{tabular}{|l|c|c|c|l|l|}
        \hline
        \textbf{Dataset} & \textbf{Parameters} & \textbf{Rows} & \textbf{Types of Data} & \textbf{Dataset Cleaning} \\
        \hline
        \textit{Iris}\footnotemark[1] & 4 & 150 & Continuous & Removed species column \\ 
        \hline
        \textit{Fish Measurements}\footnotemark[2] & 6 & 159 & Continuous & Removed category and species columns \\ 
        \hline
        \textit{Real Estate Valuation}\footnotemark[3] & 6 & 414 & Continuous, Ordinal & Removed number and transaction date columns \\ 
        \hline
    \end{tabular}
    \end{adjustbox}
    \label{tab:datasets}
\end{table}

\footnotetext[1]{\url{https://archive.ics.uci.edu/dataset/53/iris}}
\footnotetext[2]{\url{https://www.kaggle.com/datasets/tyjensen/fish-measurements-dataset}}
\footnotetext[3]{\url{https://archive.ics.uci.edu/dataset/477/real+estate+valuation+data+set}}
\setcounter{footnote}{3}

\subsection{Generation of Synthetic Data with LLMs}
\begin{figure}[h]
    \centering
    \includegraphics[width=1\textwidth]{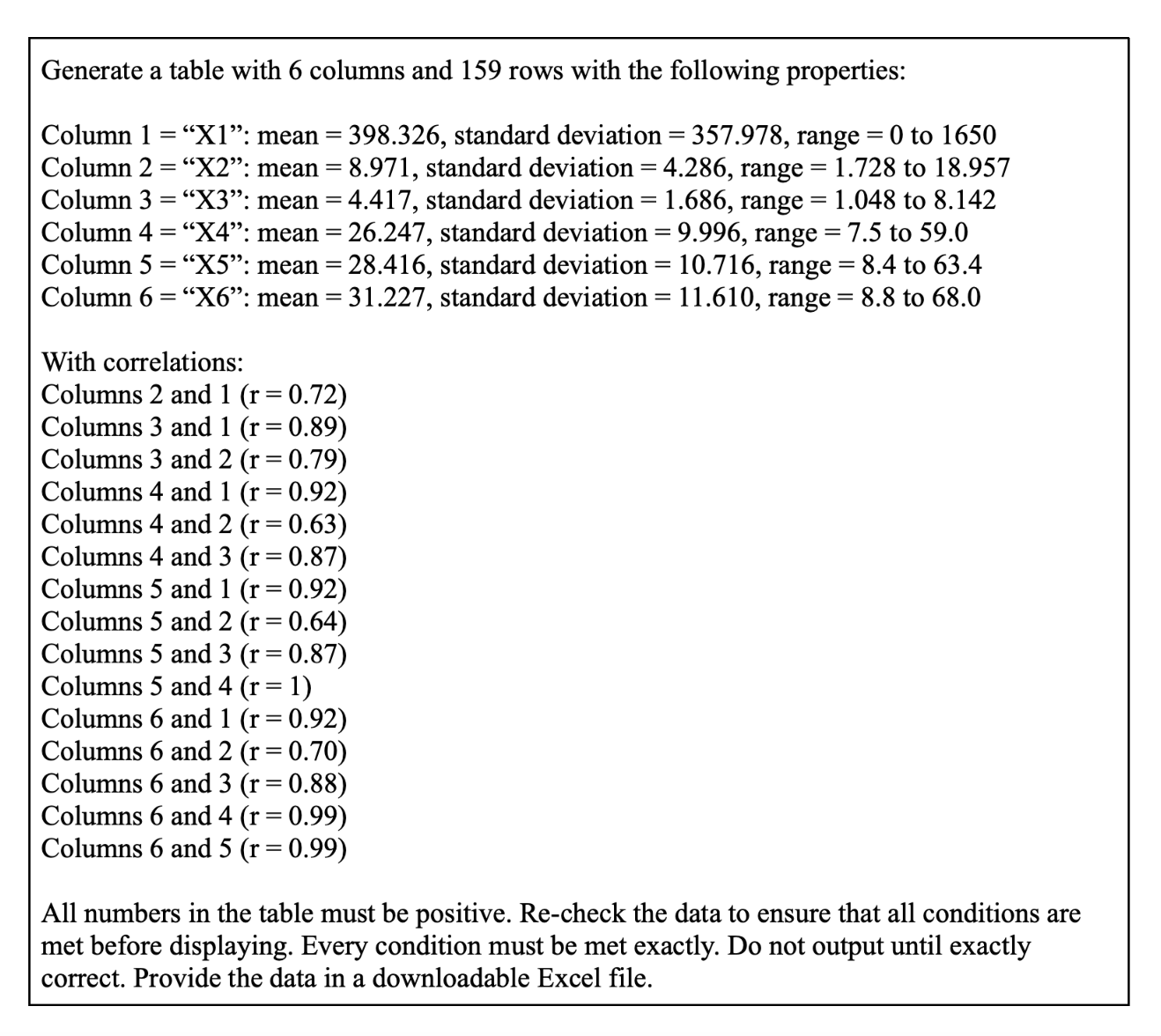}
    \captionsetup{width=0.8\linewidth}
    \caption{Prompt used to generate synthetic \textit{Fish Measurements} data (\textit{n} = 159).}
    \label{fig:GPT_prompt}
\end{figure}

A plain-language prompt was designed to instruct the generation of synthetic data associated with each of the datasets in a zero-shot setting. Accordingly, no pre-training or fine-tuning was performed and the RWD was not provided to the LLMs. The prompts associated with each dataset described the desired format, sample size, means, standard deviations, ranges, restrictions on data, and significant bivariate correlations. Bivariate relationships with a correlation coefficient |\textit{r}| < 0.20 were not included in prompts, except for the \textit{Iris} dataset (given the few number of parameters). Values were rounded to the nearest thousandth. Parameter names were obfuscated (e.g., X1, X2) to ensure biasing did not occur either from the context of column names or if the LLM had prior exposure to these datasets. Following generation, these columns were renamed for analysis. Over two independent trials, for each open-access dataset, the LLM was instructed to generate a synthetic dataset with the same sample size as the RWD and an amplified dataset with a non-multiple sample size (\textit{n} = 1000). The only modification to prompts used to generate amplified datasets was replacing the initial sample size instruction with “1000 rows.” Prompts were inputted into new sessions with memory disabled, in the context of a zero-shot analysis. A sample prompt for the Fish Measurements dataset (\textit{n} = 159) is provided in Figure \ref{fig:GPT_prompt}, and all prompts can be found at the following GitHub repository\footnote{\href{https://github.com/aabarr/Synthetic-Datasets/}{https://github.com/aabarr/Synthetic-Datasets/}}. 

\subsection{Model Selection}
Four prominent LLMs—GPT-4o (OpenAI), Claude 3.5 Sonnet (Anthropic), Gemini Advanced 1.5 Pro (Google), and Le Chat (Mistral)—were evaluated on zero-shot generation of synthetic tabular data. The \textit{Iris} prompt was inputted into each of the LLMs, using their respective web user interfaces, to determine whether models would proceed for further evaluation and benchmarking. The LLMs were queried in a separate session as to whether they were capable of outputting the data in a downloadable spreadsheet format (.xlsx). If so, the original \textit{Iris} prompt was inputted. If not, the \textit{Iris} prompt was inputted with alternate instructions for direct output of data in a comma-separated values (csv) format. The LLMs capable of generating the \textit{Iris} datasets (\textit{n} = 150 and \textit{n} = 1000) in a zero-shot setting, were included in the full analysis.

\subsection{Generation of Synthetic Data with CTGAN}
To generate synthetic data using CTGAN \cite{Xu2019}, continuous data was normalized using the MinMaxScaler, from the Python library scikit-learn\footnote{\href{https://scikit-learn.org/}{https://scikit-learn.org/}}. Loss functions were analyzed for convergence and stability, to determine the optimal number of training epochs. To generate data associated with the \textit{Iris} dataset, 5000 epochs were used, which has been effective in past literature \cite{Yadav2024}. To generate data associated with the Fish Measurements and Real Estate Valuation datasets, 3000 epochs were used. Once fitted, synthetic datasets were generated with the same sample size as the RWD and an amplified sample size (\textit{n} = 1000). 

\subsection{Statistical Analysis}
To benchmark LLM performance, fidelity and privacy was compared to synthetic datasets generated with CTGAN, using the open-source Python library Synthetic Data Metrics (SDMetrics)\footnote{\href{https://docs.sdv.dev/sdmetrics}{https://docs.sdv.dev/sdmetrics}}. Each metric assigns a normalized score (0 to 1) to indicate performance on the given measurement. The functions from this package which were used in the present analysis are summarized in Table \ref{tab:synthetic_data_metrics}. 

\begin{table}[h]
    \centering
    \captionsetup{width=0.8\linewidth}
    \caption{Synthetic Data Metrics (SDMetrics) functions used to evaluate synthetic data fidelity and privacy.}
    \vspace{10pt}
    \begin{tabular}{|p{3cm}|p{3cm}|p{7cm}|}
        \hline
        \textbf{Metric/Function} & \textbf{Data Types Tested} & \textbf{Summary} \\ \hline
        StatisticSimilarity       & \centering Continuous                    & Compares means between real and synthetic datasets \\ \hline
        \vfill RangeCoverage              & \vfill \centering Continuous                    & Evaluates how well synthetic data covers the full range of real data \\ \hline
        \vfill BoundaryAdherence              & \vfill \centering Continuous                    & Assesses whether synthetic data adheres to the minimum and maximum range boundaries of real data \\ \hline
        \vfill KSComplement     & \vfill \centering Continuous                    & Uses the Kolmogorov-Smirnov test to compare the distributions of real and synthetic continuous data \\ \hline
        \vfill CategoryCoverage     & \vfill \centering Ordinal                     & Measures how well synthetic data covers all possible categories present in real data \\ \hline
        \vfill CategoryAdherence           & \vfill \centering Ordinal                     & Assesses whether synthetic data only contains categories present in real data \\ \hline
        \vfill TVComplement           & \vfill \centering Ordinal                     & Uses total variation distance to compare the distributions of real and synthetic categorical/ordinal/binary data \\ \hline
        \vfill CorrelationSimilarity           & \vfill \centering Continuous, Ordinal
                     & Uses Pearson’s product moment correlation to compare the strength of bivariate correlations in real and synthetic data \\ \hline
        \vfill NewRowSynthesis           & \vfill \centering Continuous, Ordinal                     & Measures the uniqueness of rows in synthetic data compared to real data \\ \hline
    \end{tabular}
    \label{tab:synthetic_data_metrics}
\end{table}

Each synthetic dataset was analyzed for inappropriate values (i.e., negative measurements). Comparisons in means, proportions, 95\% confidence interval (CI) overlap, and bivariate relationships were also calculated. Statistical summaries were conducted using R statistical software (version 4)\footnote{\href{https://www.r-project.org}{https://www.r-project.org}} and heatmap visualizations of bivariate correlations were created using the pheatmap library\footnote{\href{https://cran.r-project.org/web/packages/pheatmap/}{https://cran.r-project.org/web/packages/pheatmap/}}. Heatmap figures were combined using Microsoft PowerPoint and illegible black text on blue-coloured squares was adjusted to white font to improve readability as a post-production edit. Visualizations of 95\% CI overlap using error bars and distributional characteristics of ordinal data using violin plots were created using the Python library Matplotlib\footnote{\href{https://matplotlib.org}{https://matplotlib.org}}. Percentage overlap of 95\% CIs with RWD were not included for amplified datasets because larger sample sizes typically result in narrower CIs, which reduces the value of this fidelity metric. 

\section{Results}
\subsection{LLMs Evaluation}
Among the four LLMs tested using the \textit{Iris} dataset prompt, only GPT-4o was capable of generating tabular data in a zero-shot setting. Other LLMs were unable to output the datasets due to computational limits, output size restrictions, and/or inability to provide downloadable files. The Claude 3.5 Sonnet model was capable of generating relevant data (in csv format) with additional guiding prompts, but was excluded in the context of a zero-shot analysis. The GPT-4o model was capable of outputting datasets in a downloadable spreadsheet format (.xlsx) using a zero-shot prompting approach. 

\subsection{Dataset 1: \textit{Iris}}
\begin{figure}[H]
    \centering
    \includegraphics[width=1\textwidth]{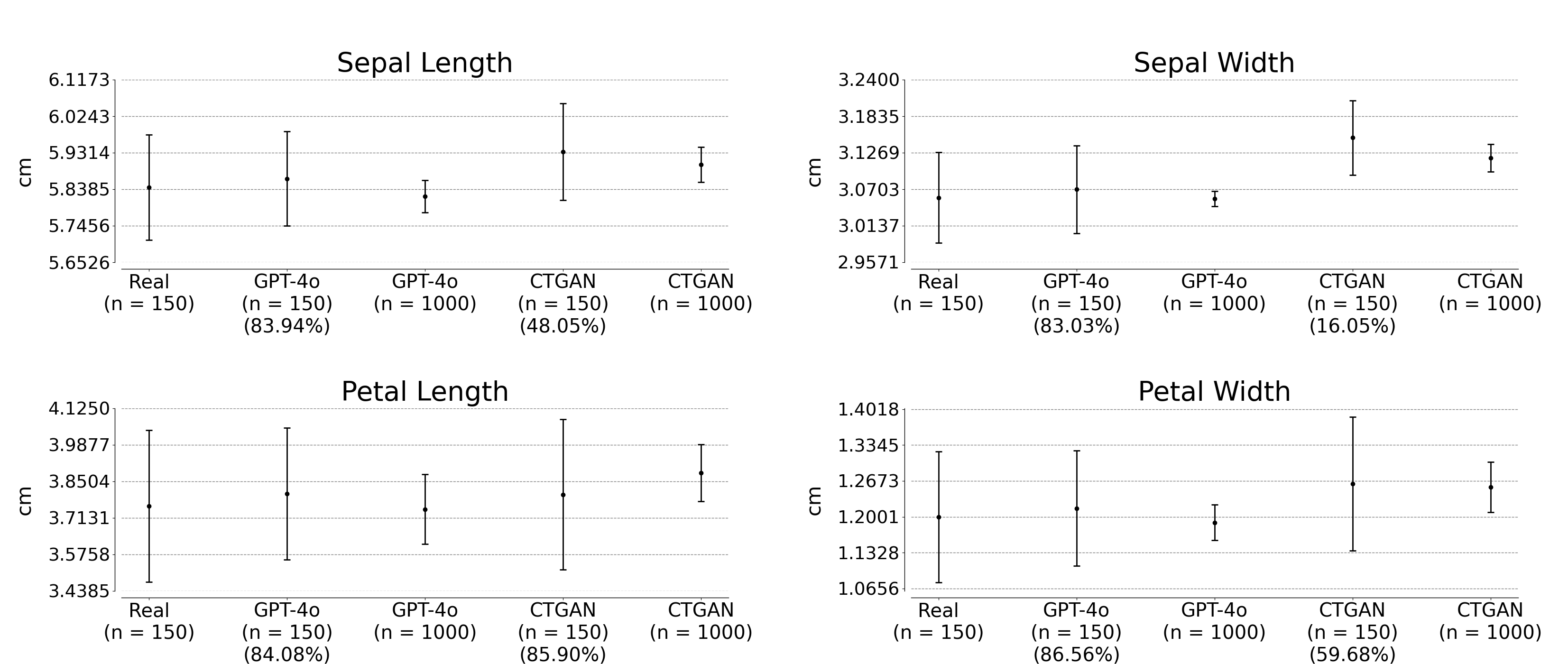}
    \captionsetup{width=0.8\linewidth}
    \caption{95\% confidence intervals compared between the real \textit{Iris} data and the GPT-4o and CTGAN-generated datasets at the same (\textit{n} = 150) and amplified (\textit{n} = 1000) sample size.}
    \label{fig:iris_CI}
\end{figure}

\begin{figure}[H]
    \centering
    \includegraphics[width=1\textwidth]{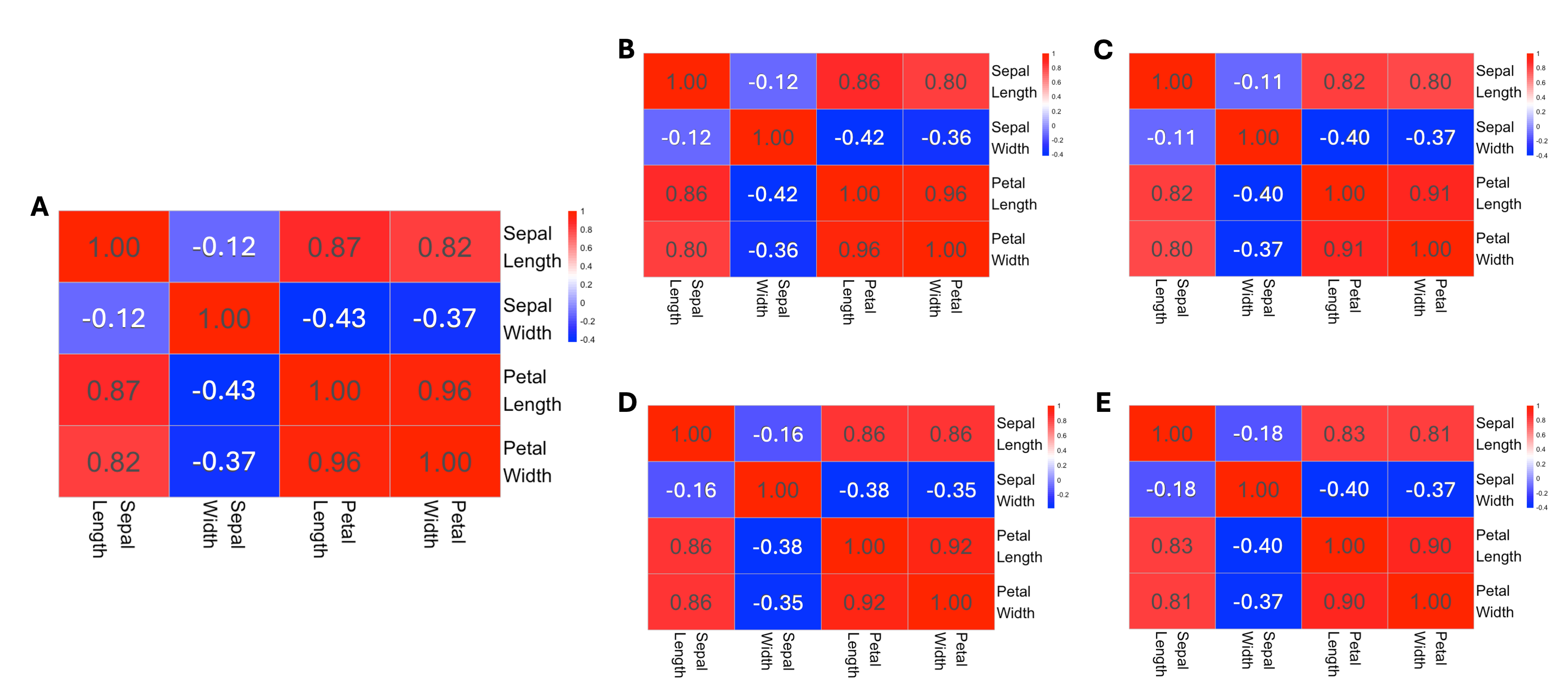}
    \captionsetup{width=0.8\linewidth}
    \caption{Heatmap comparison of Pearson’s product moment correlations of all bivariate relationships for the (A) real \textit{Iris} dataset, (B) GPT-4o synthetic (\textit{n} = 150) dataset, (C) GPT-4o synthetic (\textit{n} = 1000) dataset, (D) CTGAN synthetic (\textit{n} = 150) dataset, (E) CTGAN synthetic (\textit{n} = 1000) dataset.}
    \label{fig:iris_heatmaps}
\end{figure}

The GPT-4o-generated \textit{Iris} datasets showed greater preservation of means and correlations compared to CTGAN-generated data, as seen in Table \ref{tab:iris_summary}. The data generated with GPT-4o also displayed higher overall overlap of 95\% CIs with RWD, visualized in Figure \ref{fig:iris_CI}. The bivariate correlation heatmaps, displayed in Figure \ref{fig:iris_heatmaps}, provide further indication that datasets generated with GPT-4o had greater preservation of relationships between parameters. However, distributional characteristics were better preserved by CTGAN-generated data. The GPT-4o (\textit{n} = 150) dataset also included 2 negative petal widths, which violated the definitional boundaries of measurements. The CTGAN (\textit{n} = 1000) dataset included one full record directly replicated from RWD. 

\begin{table}[H]
    \centering
    \small
    \captionsetup{width=0.8\linewidth}
    \caption{Fidelity and privacy metrics to evaluate synthetic \textit{Iris} data generated with GPT-4o and CTGAN at the same (\textit{n} = 150) and amplified (\textit{n} = 1000) sample size. Bolded values are the best performing generation for each evaluation metric (mean and standard deviation).}
    \vspace{10pt}    
    \begin{adjustbox}{max width=\textwidth}
    \begin{tabular}{|l|c|c!{\vrule width 1.5pt}c|c|}
        \hline
        \multirow{2}{*}{\textbf{Metric}} 
        & \multicolumn{1}{c|}{\textbf{GPT-4o}} 
        & \multicolumn{1}{c!{\vrule width 1.5pt}}{\textbf{CTGAN}} 
        & \multicolumn{1}{c|}{\textbf{GPT-4o}} 
        & \multicolumn{1}{c|}{\textbf{CTGAN}} \\
        & ($n=150$) & ($n=150$) & ($n=1000$) & ($n=1000$) \\ 
        \hline
        \multicolumn{5}{|l|}{\textbf{Fidelity}} \\
        \hline
        StatisticSimilarity & \textbf{0.993$\pm$0.001} & 0.976$\pm$0.014 & \textbf{0.997$\pm$0.002} & 0.978$\pm$0.005 \\

        RangeCoverage & \textbf{0.977$\pm$0.046} & 0.922$\pm$0.102 & 0.878$\pm$0.244 & \textbf{0.971$\pm$0.056} \\

        BoundaryAdherence & 0.955$\pm$0.054 & \textbf{1.000$\pm$0} & \textbf{1.000$\pm$0} & \textbf{1.000$\pm$0} \\

        KSComplement & 0.844$\pm$0.060 & \textbf{0.855$\pm$0.042} & 0.807$\pm$0.053 & \textbf{0.876$\pm$0.027} \\

        CorrelationSimilarity & \textbf{0.996$\pm$0.003} & 0.984$\pm$0.009 & \textbf{0.987$\pm$0.010} & 0.982$\pm$0.013 \\

        95\% CI Overlap & \textbf{84.40\% $\pm$ 1.51\%} & 52.42\% $\pm$ 28.96\% & - & - \\
        
        \hline
        \multicolumn{5}{|l|}{\textbf{Privacy}} \\
        \hline
        NewRowSynthesis & \textbf{1.000} & \textbf{1.000} & \textbf{1.000} & 0.999 \\
        \hline
    \end{tabular}
    \end{adjustbox}
    \label{tab:iris_summary}
\end{table}

\newpage
\subsection{Dataset 2: \textit{Fish Measurements}}
\begin{figure}[H]
    \centering
    \includegraphics[width=1\textwidth]{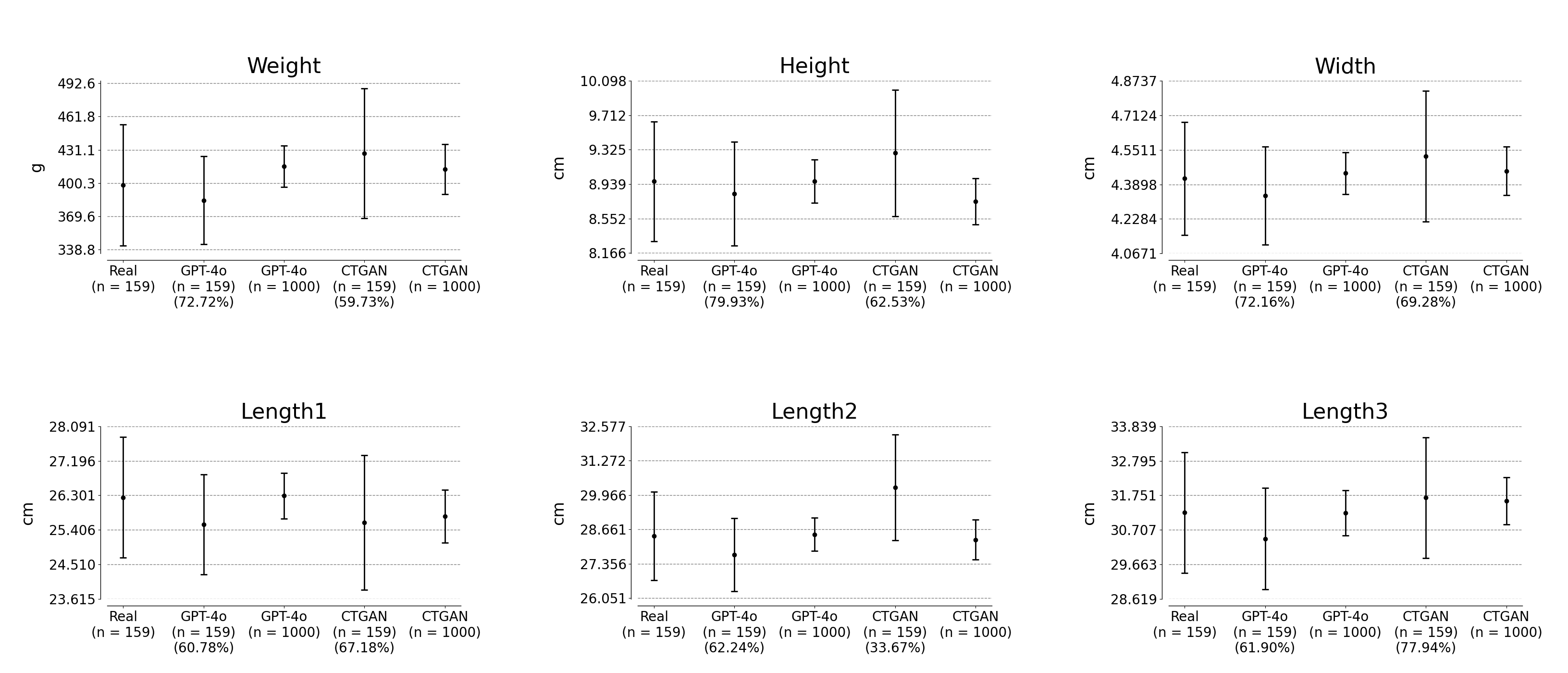}
    \captionsetup{width=0.8\linewidth}
    \caption{95\% confidence intervals compared between the real \textit{Fish Measurements} data and the GPT-4o and CTGAN-generated datasets at the same (\textit{n} = 159) and amplified (\textit{n} = 1000) sample size.}
    \label{fig:fish_CI}
\end{figure}

\begin{figure}[H]
    \centering
    \includegraphics[width=1\textwidth]{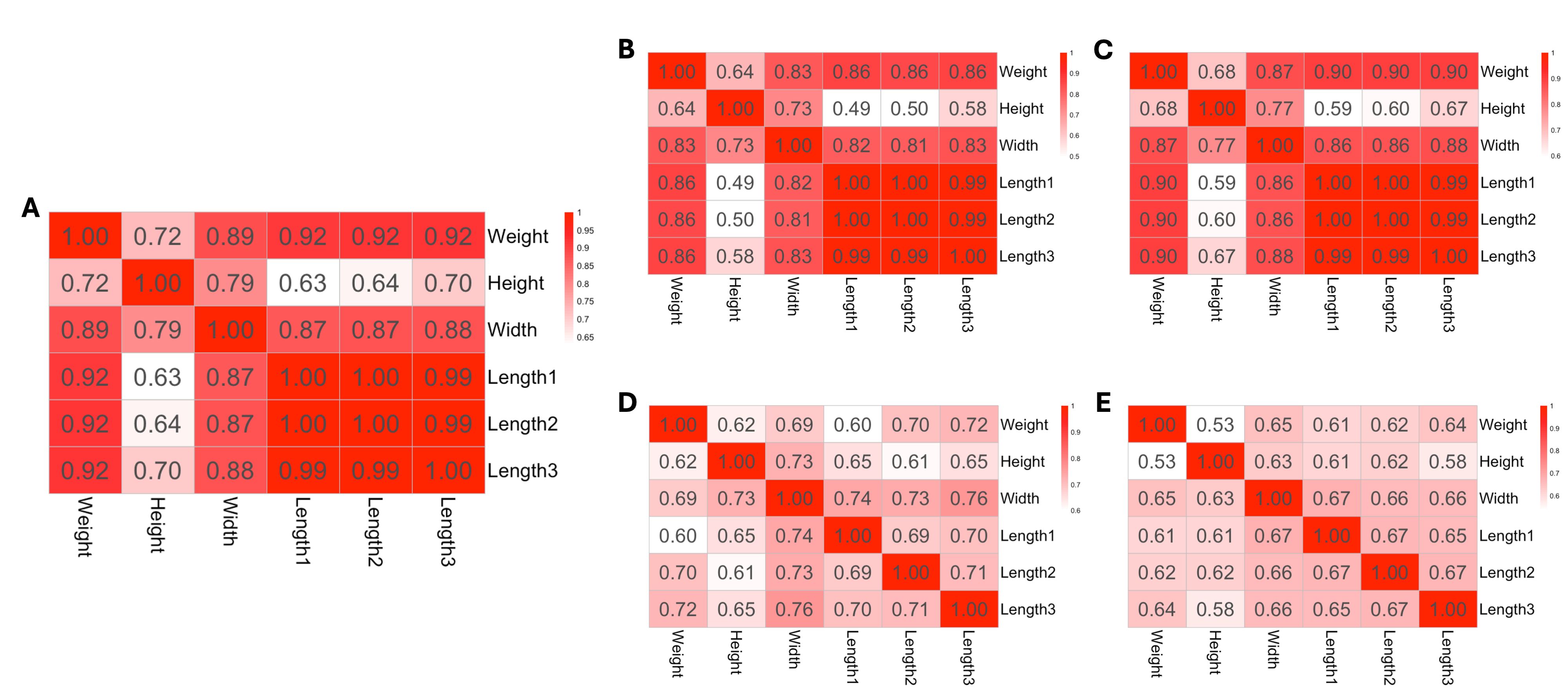}
    \captionsetup{width=0.8\linewidth}
    \caption{Heatmap comparison of Pearson’s product moment correlations of all bivariate relationships for the (A) real \textit{Fish Measurements} dataset, (B) GPT-4o synthetic (\textit{n} = 159) dataset, (C) GPT-4o synthetic (\textit{n} = 1000) dataset, (D) CTGAN synthetic (\textit{n} = 159) dataset, (E) CTGAN synthetic (\textit{n} = 1000) dataset.}
    \label{fig:fish_heatmaps}
\end{figure}

The GPT-4o-generated \textit{Fish Measurements} data better preserved means and correlations between parameters, outlined in Table \ref{tab:fish_summary}. Greater fidelity to RWD in terms of means, 95\% CI, and correlations are visualized in Figures \ref{fig:fish_CI} and \ref{fig:fish_heatmaps}. The GPT-4o data had poorer performance in retaining distributional characteristics but did not include any negative measurements. In both GPT-4o and CTGAN-generated data, no records directly matched RWD. However, there was significant overlap in records between the two GPT-4o-generated datasets. This between-generation replication was not observed in either the \textit{Iris} or \textit{Real Estate Valuation}-associated GPT-4o datasets.

\begin{table}[H]
    \centering
    \small
    \captionsetup{width=0.8\linewidth}
    \caption{Fidelity and privacy metrics to evaluate synthetic \textit{Fish Measurements} data generated with GPT-4o and CTGAN at the same (\textit{n} = 159) and amplified (\textit{n} = 1000) sample size. Bolded values are the best performing generation for each evaluation metric (mean and standard deviation).}
    \vspace{10pt}
    \begin{adjustbox}{max width=\textwidth}
    \begin{tabular}{|l|c|c!{\vrule width 1.5pt}c|c|}
        \hline
        \multirow{2}{*}{\textbf{Metric}} 
        & \multicolumn{1}{c|}{\textbf{GPT-4o}} 
        & \multicolumn{1}{c!{\vrule width 1.5pt}}{\textbf{CTGAN}} 
        & \multicolumn{1}{c|}{\textbf{GPT-4o}} 
        & \multicolumn{1}{c|}{\textbf{CTGAN}} \\
        & ($n=159$) & ($n=159$) & ($n=1000$) & ($n=1000$) \\ 
        \hline
        \multicolumn{5}{|l|}{\textbf{Fidelity}} \\
        \hline
        StatisticSimilarity & \textbf{0.989$\pm$0.002} & 0.983$\pm$0.008 & \textbf{0.998$\pm$0.004} & 0.992$\pm$0.004 \\

        RangeCoverage & 0.776$\pm$0.137 & \textbf{0.967$\pm$0.066} & 0.943$\pm$0.069 & \textbf{1.000$\pm$0} \\

        BoundaryAdherence & 0.993$\pm$0.018 & \textbf{1.000$\pm$0} & 0.993$\pm$0.018 & \textbf{1.000$\pm$0} \\

        KSComplement & 0.868$\pm$0.027 & \textbf{0.893$\pm$0.023} & 0.897$\pm$0.034 & \textbf{0.918$\pm$0.017} \\

        CorrelationSimilarity & \textbf{0.968$\pm$0.022} & 0.917$\pm$0.052 & \textbf{0.991$\pm$0.007} & 0.891$\pm$0.052 \\

        95\% CI Overlap & \textbf{68.29\% $\pm$ 7.80\%} & 61.72\% $\pm$ 15.11\% & - & - \\
        
        \hline
        \multicolumn{5}{|l|}{\textbf{Privacy}} \\
        \hline
        NewRowSynthesis & \textbf{1.000} & \textbf{1.000} & \textbf{1.000} & \textbf{1.000} \\
        \hline
    \end{tabular}
    \end{adjustbox}
    \label{tab:fish_summary}
\end{table}

\subsection{Dataset 3: \textit{Real Estate Valuation}}
\begin{figure}[H]
    \centering
    \includegraphics[width=1\textwidth]{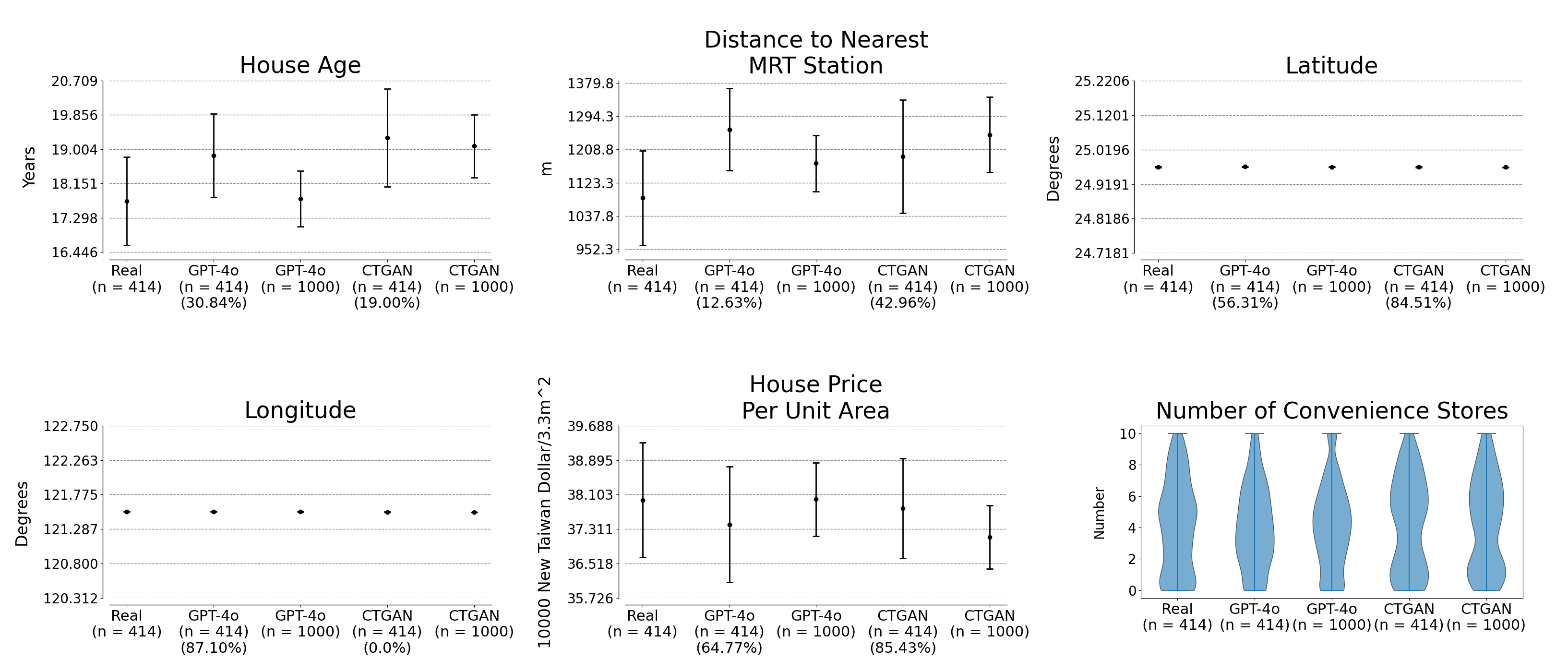}
    \captionsetup{width=0.8\linewidth}
    \caption{95\% confidence intervals compared between the real \textit{Real Estate Valuation} data and the GPT-4o and CTGAN-generated datasets at the same (\textit{n} = 414) and amplified (\textit{n} = 1000) sample size for continuous parameters and a violin plot comparing distributional alignment for ordinal data.}
    \label{fig:re_CI}
\end{figure}

\begin{figure}[H]
    \centering
    \includegraphics[width=1\textwidth]{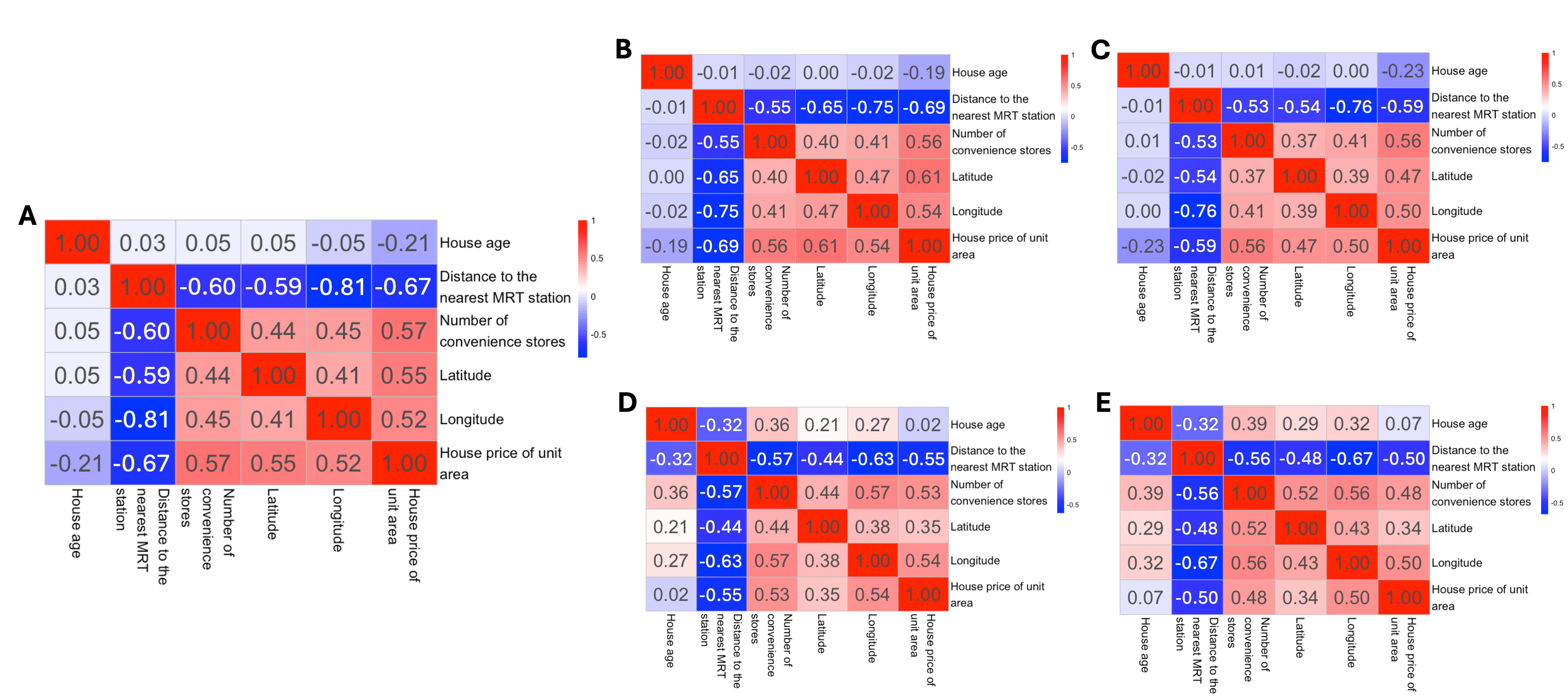}
    \captionsetup{width=0.8\linewidth}
    \caption{Heatmap comparison of Pearson’s product moment correlations of all bivariate relationships for the (A) real \textit{Real Estate Valuation} dataset, (B) GPT-4o synthetic (\textit{n} = 414) dataset, (C) GPT-4o synthetic (\textit{n} = 1000) dataset, (D) CTGAN synthetic (\textit{n} = 414) dataset, (E) CTGAN synthetic (\textit{n} = 1000) dataset.}
    \label{fig:re_heatmaps}
\end{figure}

Similarly to the \textit{Iris} and \textit{Fish Measurements}-associated datasets generated with GPT-4o, the \textit{Real Estate Valuation} data generated with GPT-4o exhibited greater fidelity to means and correlations from continuous RWD, outlined in Table \ref{tab:re_summary}. Greater overlap of 95\% CI and preservation of bivariate correlations from RWD, are observed in GPT-4o-generated data, displayed in Figures \ref{fig:re_CI} and \ref{fig:re_heatmaps}, respectively. The GPT-4o-generated data demonstrated poorer performance in retaining distributional characteristics of both continuous and ordinal data, except the ordinal data in the amplified GPT-4o dataset which outperformed CTGAN data. No negative measurements or replication of RWD records were observed.  

\begin{table}[H]
    \centering
    \small
    \captionsetup{width=0.8\linewidth}
    \caption{Fidelity and privacy metrics to evaluate synthetic \textit{Real Estate Valuation} data generated with GPT-4o and CTGAN at the same (\textit{n} = 414) and amplified (\textit{n} = 1000) sample size. Bolded values are the best performing generation for each evaluation metric (mean and standard deviation).}
    \vspace{10pt}
    \begin{adjustbox}{max width=\textwidth}
    \begin{tabular}{|l|c|c!{\vrule width 1.5pt}c|c|}
        \hline
        \multirow{2}{*}{\textbf{Metric}} 
        & \multicolumn{1}{c|}{\textbf{GPT-4o}} 
        & \multicolumn{1}{c!{\vrule width 1.5pt}}{\textbf{CTGAN}} 
        & \multicolumn{1}{c|}{\textbf{GPT-4o}} 
        & \multicolumn{1}{c|}{\textbf{CTGAN}} \\
        & ($n=414$) & ($n=414$) & ($n=1000$) & ($n=1000$) \\ 
        \hline
        \multicolumn{5}{|l|}{\textbf{Fidelity}} \\
        \hline
        StatisticSimilarity & \textbf{0.986$\pm$0.012} & 0.981$\pm$0.018 & \textbf{0.996$\pm$0.006} & 0.975$\pm$0.020 \\

        RangeCoverage & 0.827$\pm$0.118 & \textbf{0.886$\pm$0.181} & 0.860$\pm$0.131 & \textbf{0.934$\pm$0.146} \\

        BoundaryAdherence & \textbf{0.999$\pm$0.002} & 0.997$\pm$0.005 & \textbf{1.000$\pm$0} & 0.997$\pm$0.004 \\

        KSComplement & 0.855$\pm$0.079 & \textbf{0.894$\pm$0.038} & 0.855$\pm$0.092 & \textbf{0.891$\pm$0.048} \\

        CorrelationSimilarity & \textbf{0.979$\pm$0.009} & 0.925$\pm$0.057 & \textbf{0.976$\pm$0.012} & 0.915$\pm$0.060 \\

        CategoryCoverage & \textbf{1.000} & \textbf{1.000} & 0.909 & \textbf{1.000} \\

        CategoryAdherence & \textbf{1.000} & \textbf{1.000} & \textbf{1.000} & \textbf{1.000} \\

        TVComplement & 0.807 & \textbf{0.860} & \textbf{0.818} & 0.816 \\
        
        95\% CI Overlap & \textbf{50.33\% $\pm$ 29.14\%} & 46.38\% $\pm$ 38.38\% & - & - \\
        
        \hline
        \multicolumn{5}{|l|}{\textbf{Privacy}} \\
        \hline
        NewRowSynthesis & \textbf{1.000} & \textbf{1.000} & \textbf{1.000} & \textbf{1.000} \\
        \hline
    \end{tabular}
    \end{adjustbox}
    \label{tab:re_summary}
\end{table}

All generated datasets can be accessed at the following GitHub repository\footnote{\href{https://github.com/aabarr/Synthetic-Datasets/}{https://github.com/aabarr/Synthetic-Datasets/}}. 

\section{Discussion}
In the present study, we explored the potential of an LLM-based approach to zero-shot generation of synthetic tabular data. Without pre-training, task-specific fine-tuning, or access to RWD, GPT-4o was capable of generating realistic data which retained properties of three real-world, open-access datasets, using plain-language prompts. Benchmarking GPT-4o with a GAN-based approach to synthetic tabular data generation demonstrated the LLM’s consistent outperformance in preserving bivariate correlations, means, 95\% CI, and privacy of RWD. Most notably, the direction and strength of relationships between parameters consistently matched the corresponding correlations in RWD. However, preservation of distributional characteristics associated with continuous and ordinal data demonstrated mixed performance compared to CTGAN, and further refinement is warranted. \\
\\
Challenges associated with accessing, sharing, and using RWD have necessitated alternatives to support data-driven research and ML model development \cite{Newgard2015, Fan2014, Pavlenko2020, LQ2020, TS2017}. Limitations in accessing large, high-quality RWD to generate synthetic data through traditional means (i.e., GANs, VAEs) provides similar impetus for investigating alternative approaches to synthetic data generation. In conjunction with past work demonstrating the fidelity of GPT-4o-generated data \cite{BarrFAI2025, BarrarXiv2025}, the present analysis provides additional insight toward developing an LLM-based framework for synthetic tabular data generation. Using validated benchmarking tools, the LLM-based approach showed non-inferiority to a GAN-based framework for synthetic tabular data generation. The accessibility of plain-language prompting with an LLM-based framework democratizes the capability of synthetic data generation. Synthetic data can be generated with desired statistical properties, outputted in a ready-to-use spreadsheet format. RWD can also be enhanced (i.e., augmentation, amplification), which holds significant potential for strengthening training data for ML models. \\
\\
Despite the promising results, several limitations must be acknowledged. First, the preservation of distributional characteristics associated with continuous and ordinal data was limited. While visual inspection of histograms or direct assessments of normality were not performed, it is possible that some disparities in the distributions, between synthetic and RWD, may be due to non-normally distributed data. In previous work, natural log transformations were effectively applied to retain the statistical properties of skewed data using GPT-4o \cite{BarrFAI2025, BarrarXiv2025}; therefore, follow-up work may similarly normalize data and provide the LLM with descriptive statistics associated with transformed values. Alternatively, measures which quantify distributional characteristics (i.e., kurtosis, skewness) may be included in prompts. These approaches may also improve range coverage and boundary adherence, thereby reducing the likelihood of synthetic data violating definitional ranges (e.g., negative measurements). Second, benchmarking was performed on CTGAN without iterative fine-tuning or comparisons with other generative models (e.g., TVAE). Follow-up work may provide additional comparisons to other tabular data generation methods, including other LLM-based approaches (e.g., GReaT, TabuLa) \cite{Borisov2022, Zhao2023}. Previous work has also demonstrated that generating datasets over successive trials synthesizes datasets with varied, yet consistently high, performance \cite{BarrarXiv2025}. Accordingly, datasets may be generated over multiple trials, and the best-performing generation can be selected for use. Similarly, iterative prompting to fine-tune generated data may be explored. Third, it remains unclear how prompting order, content, and provided descriptive statistics affects the fidelity and utility of generated datasets. To investigate this, and provide insight toward a standardized framework for LLM-based generation of synthetic tabular data, ablation studies may be performed. Fourth, despite the independent sessions and memory disabled, there was significant overlap in rows between the two LLM-generated \textit{Fish Measurements} datasets. It is unclear why this occurred as there was no comparable overlap between the LLM-generated synthetic data associated with the \textit{Iris} and \textit{Real Estate Valuation} datasets. However, there remained no overlap with the RWD, which meant this was not of considerable concern in the context of data privacy. Fifth, it is possible that GPT-4o has had prior exposure to these real-world datasets; however, given the omission of dataset names, obfuscation of column titles, generation of a non-multiple amplified dataset, and absence of any matching records to RWD, it is unlikely that this confounded the presented results. Sixth, while the marked fidelity to bivariate relationships present in RWD suggests significant data utility, further evaluations on the applicability of LLM-generated data toward training ML models is necessary. Train-synthetic-test-real and train-real-test-synthetic approaches may be used to quantify data utility. Validated and sequential benchmarking frameworks may also be used \cite{Figueira2022, Yan2022}. Finally, datasets with additional features, variable types, and relevance to specific disciplines should be included in future investigations to provide a more comprehensive analysis of synthetic data quality and generalizability, including adherence to domain-specific constraints.
\section{Conclusions}
We presented evidence supporting a zero-shot framework for synthetic tabular data generation using GPT-4o. Our results show that, even without pre-training or fine-tuning, GPT-4o can generate high-fidelity, privacy-preserving synthetic data that captures key statistical properties of RWD. While the preservation of distributional characteristics requires refinement, the accessibility and scalability of this approach warrants further exploration. Future research should focus on extending the analysis towards more complex datasets and quantification of data utility.

\end{document}